\documentclass[runningheads]{llncs}
\usepackage[T1]{fontenc}
\usepackage{graphicx}
\usepackage{booktabs}
\usepackage{algorithm}
\usepackage{algorithmic}
\usepackage{amssymb}
\usepackage{multirow}
\usepackage[section]{placeins}
\usepackage{bbding}
\usepackage{subfigure}
\usepackage{wrapfig}
\usepackage{amsmath}
\usepackage[marginal]{footmisc}

\begin{document}
\title{FedDTG:Federated Data-Free Knowledge Distillation via Three-Player Generative Adversarial Networks}
\titlerunning{Federated Data-Free Distillation via Three-Player GAN}
\author{Lingzhi Gao\inst{1\footnotemark[1]} \and
Zhenyuan Zhang\inst{2\footnotemark[1]} \and
Chao Wu\inst{1(}\Envelope\inst{)}}

\authorrunning{L. Gao et al.}
\institute{
Zhejiang University, Hangzhou, China \\
% \email{\{lingzhigao,22051061,chao.wu\}@zju.edu.cn}
\and
Zhejiang Rural Commercial United Bank Co.,Ltd., Hangzhou, China
}
\maketitle              % typeset the header of the contribution
\renewcommand{\thefootnote}{\fnsymbol{footnote}}
\footnotetext[1]{First Author and Second Author contribute equally to this work.\\}
\begin{abstract}
While existing federated learning approaches primarily focus on aggregating local models to construct a global model, in realistic settings, some clients may be reluctant to share their private models due to the inclusion of privacy-sensitive information.
Knowledge distillation, which can extract model knowledge without accessing model parameters, is well-suited for this federated scenario. 
However, most distillation methods in federated learning (federated distillation) require a proxy dataset, which is difficult to obtain in the real world.  Therefore, in this paper, we introduce a distributed three-player Generative Adversarial Network (GAN) to implement data-free mutual distillation and propose an effective method called FedDTG. We confirmed that the fake samples generated by GAN can make federated distillation more efficient and robust. 
Additionally, the distillation process between clients can deliver good individual client performance while simultaneously acquiring global knowledge and protecting data privacy.
Our extensive experiments on benchmark vision datasets demonstrate that our method outperforms other federated distillation algorithms in terms of generalization.

% \keywords{Federated learning  \and Knowledge distillation \and Generative adversarial networks.}
\end{abstract}
\section{Introduction}
Federated Learning (FL) \cite{Fedsurvey,fedavg} is a decentralized machine learning setting, in which multiple clients participate in collaborative training under the coordination of a central server, without the need to share their private data. 
Different from traditional centralized learning, FL protects clients' data privacy well, while utilizing data from edge nodes, thereby addressing the data silo problem\cite{huang2021personalized,Fedsurvey}.

However, in real-world scenarios, the data from each client varies significantly due to differences in their lifestyles and behaviors, leading to non-independent identical distribution (non-iid) data. This data heterogeneity \cite{noniid} poses a tough challenge in FL, severely impacting the model's convergence speed during training and potentially causing the model to optimize in the wrong direction\cite{Fedsurvey,fedprox}.

Many approaches to addressing the data heterogeneity in federated learning can be categorized into client-side methods and server-side methods. The traditional FedAvg \cite{fedavg} algorithm averages the parameters of all client models to get a global model. Building upon this foundation, client-side methods usually improve local updates by stabilizing local training on the client before transmitting the updated parameters to the server for averaging \cite{scaffold,fedprox}. 
However, models uploaded by clients are vulnerable to inference attacks in practical applications, and adversaries may exploit model parameters to restore the local data \cite{yin2021see}. Although the availability of encryption techniques like differential privacy\cite{dwork2006differential}, this risk discourages some clients from uploading their models. Simultaneously, the model structures may differ among clients. These issues make it difficult to directly average and aggregate local model parameters.

Sever-side methods, represented by federated distillation \cite{fedmd,fedgen}, applying knowledge distillation (KD) \cite{hinton2015distilling} to FL. These approaches enable the extraction of knowledge from other clients without accessing their model parameters. Instead, clients only share their soft labels on a proxy dataset for subsequent distillation which alleviates the model structure drift issue. However, the effectiveness of these approaches relies on the availability of a proxy dataset that is relevant to the clients' data. The consistency in distribution between the proxy dataset and the global dataset, as well as their alignment within the same domain, significantly impacts the performance of federated distillation. In some scenarios like financial applications\cite{long2020federated} or healthcare\cite{sheller2020federated}, where it is difficult to collect data, the assumption regarding the availability of a suitable proxy dataset may be unrealistic.

Observing the limitations of the previous work, we propose FedDTG (\textbf{Fed}erated \textbf{D}istillation via a \textbf{T}hree-player \textbf{G}enerative Adversarial Networks). FedDTG uses a distributed three-player GAN to generate fake samples for mutual distillation \cite{shen2020federated,zhang2018deep} between clients, which can be trained simultaneously with federated distillation and local personalized tasks. The generated synthetic samples can not only regularize local training by augmenting the missing local data but also enhance the efficiency and robustness of federated distillation. Moreover, FedDTG allows each client to have a different model structure. In addition, when we need to do various tasks on similar datasets, we do not need to retrain FedDTG. The samples generated by FedDTG can be directly applied to all federated distillation algorithms. 

To sum up, our main contributions are:
    \begin{itemize}
        \item We propose a federated distillation approach FedDTG that does not require a proxy dataset. Through mutual distillation between clients, FedDTG achieves superior performance compared to strong federated distillation algorithms. 
        \item Our FedDTG learns a global generative model, which can be used to augment local datasets to make the algorithm more stable and converge faster.

        \item We validate the effectiveness of FedDTG across diverse image datasets characterized by varying extents of data heterogeneity. Our results prove that our method effectively mitigates differences in accuracy among clients, exhibits robustness, and achieves superior efficiency.
        
    \end{itemize}

\section{Related Work}
\subsection{Federated Learning}
As a classic algorithm in Federated Learning (FL), FedAvg \cite{fedavg} allows clients to conduct local training and subsequently forward the local model weights to a server. The server then performs weighted averaging on these local weights, proportional to the local training size, to produce a global model. However, this method does not perform well with non-iid data. To improve it, there have been many studies trying to stabilize local training. FedProx \cite{fedprox} proposes a proximal term to improve local training. By adding a proximal term to the local objective, the update of the local model can be restricted by the global model. Similarly, SCAFFOLD \cite{scaffold} uses the difference between the local control variates and the global control variates to correct the local updates. However, the generalization of these methods remains to be verified. As shown in the \cite{li2021model}, those studies offer minimal or no improvement over FedAvg under certain non-iid settings.
\subsection{Federated Distillation}
Knowledge distillation is initially introduced to extract knowledge from teacher models and condense it into a small student model. In \cite{hinton2015distilling}, the student model learns by emulating the teacher model's knowledge to gain its capabilities, and the knowledge is defined as softened logits. Most methods of knowledge distillation require only the outputs of the hidden layer or output layer. Compared to directly exchanging the model parameters, federated distillation only requires lower communication costs to achieve better aggregation results. FedMD \cite{fedmd} uses transfer learning to combine knowledge distillation and federated learning. In this approach, each client participating in this communication round calculates its class scores on the public dataset and transmits the result to a central server. Then the server averages the received class scores to get the updated consensus. This consensus represents the knowledge of all participating client models and each client trains the model to align with the consensus derived from the public dataset. In this way, each client model can learn the knowledge of other client models without sharing its private data or models. However, this method brings a loss of accuracy and huge differences between clients in the case of heterogeneous users. FedDF \cite{feddf}  improves the efficacy of model aggregation. They use the global model in FedAvg as the student model and do ensemble distillation to get the knowledge from all client teacher models. The ensemble knowledge is represented by the average logit outputs of all parties on an unlabeled dataset from other domains. However only refining the global model can not completely take advantage of distillation to solve non-iid problems, which may even weaken the effect of knowledge distillation. FedGen \cite{fedgen} is the first to combine data-free distillation with federated learning. FedGen only needs to share a lightweight generator model and the prediction layer of the local model with the server for averaging. The global generator outputs feature representations, which are the input of the prediction layer, to reinforce the local learning with a better generalization performance. However, due to the over-reliance on the global generator, which conveys all ensemble client information, it fails to leverage the advantages of knowledge distillation fully.
\subsection{Federated GANs}
Generative adversarial networks (GAN) \cite{goodfellow2014generative} is a generative model, which provides a way to learn deep representations without an extensively annotated training dataset. Distributed GANs are proposed to train a generator over datasets from multiple clients. MD-GAN \cite{mdgan} consists of a 
single global generator and distributed discriminators. The global generator plays a game with the ensemble of all participant-distributed Discriminators. The server in MD-GAN sends two distinct batches, which are composed of the data generated by the global generator, to each party and the client performs some learning iterations on its discriminator. Then the client sends the error feedback to the server. Finally, the server computers the average of all feedback to update its parameters. During the training process, clients iteratively exchange their discriminators with each other to avoid overfitting. With similar settings, the global generator in \cite{yonetani2019decentralized} only needs to fool the weakest individual discriminator. However, the communication cost caused by the transmission of the generated data is too high. Some works directly apply FedAvg to generators and discriminators \cite{g3,g1,g2}. These GANs in Federated learning can provide proxy data for federated distillation algorithms, but such separate training adds a great deal of computational cost. Our method adds a classifier for a three-player game. The classifier can help the generator learn the data distribution faster and also serve as a personalized task for the client.

\section{Method}
% \subsection{Notations}

% \begin{table}[h]
%         \centering
%         \caption{Table of notations.}
%         \begin{tabular}{c|c}
%             \hline \textbf{Notation}&\textbf{Description}\\
%             \hline $\mathbf{G}$&Generator\\
%             \hline $\mathbf{D}$&Discriminator\\
%             \hline $\mathbf{C}$&Classifier\\
%             \hline $N$&Total number of clients\\
%             \hline $M$&Number of active clients per round\\
%             \hline $n$&Total number of image class\\
%             \hline $r$&Sampling ratio of training dataset\\
%             \hline $data_k$&Local data k\\
%             \hline $p_{data}$&Global data distribution\\
%             \hline $p_{\mathbf{G}}$&Generated data distribution\\
%             \hline
%         \end{tabular}
%         \label{t1}
% \end{table}
    
Our work is conducted in the traditional FL setting. Each local client $Local_k$ has a Generator $\mathbf{G}_k$, a Discriminator $\mathbf{D}_k$, and a Classifier $\mathbf{C}_k$ which represents the local personalized training task. $Local_k$ owns a labeled private dataset $data_k$. Let $p_{data}$ be the global true data distribution, which cannot be observed by conventional FL, and $p_{\mathbf{G}}$ be the generated sample distribution.

The goal for FedDTG is to train $N$ different classifiers that perform well on the corresponding client tasks and a global generator $\mathbf{G}$ to make $p_{data}$ and $p_{\mathbf{G}}$ as similar as possible. The training process can be divided into three steps:

\noindent \textbf{Local Adversarial Training:} Each client trains local models on the private $data_k$ and generated samples for a few epochs, and then sends the $\mathbf{G}_k$ and $\mathbf{D}_k$ to the server.

\noindent \textbf{Global Aggregation:} The server aggregates the received generators and discriminators to get the global $\mathbf{G}$ and $\mathbf{D}$ and send it back to the client.

\noindent \textbf{Federated Distillation:} Each client inputs fake samples, which are generated by the received global $\mathbf{G}$, into $\mathbf{C}_k$ to get the soft label and sends it to the server. The server calculates the average value of other soft labels and returns it. Finally, each client uses this average soft label to calculate the loss of knowledge distillation and updates $\mathbf{C}_k$. The transmission of information between clients in FedDTG mainly comes from two parts: The first part is the global generator $\mathbf{G}$, including the correct classification of the generated images in the Local Adversarial Training stage and the Federated Distillation stage. Another part is the average of the soft labels of other clients received from the server, that is, the mutual distillation in the Federated Distillation stage.

\begin{figure}[h]
    \centering
    \includegraphics[width=1.0\textwidth]{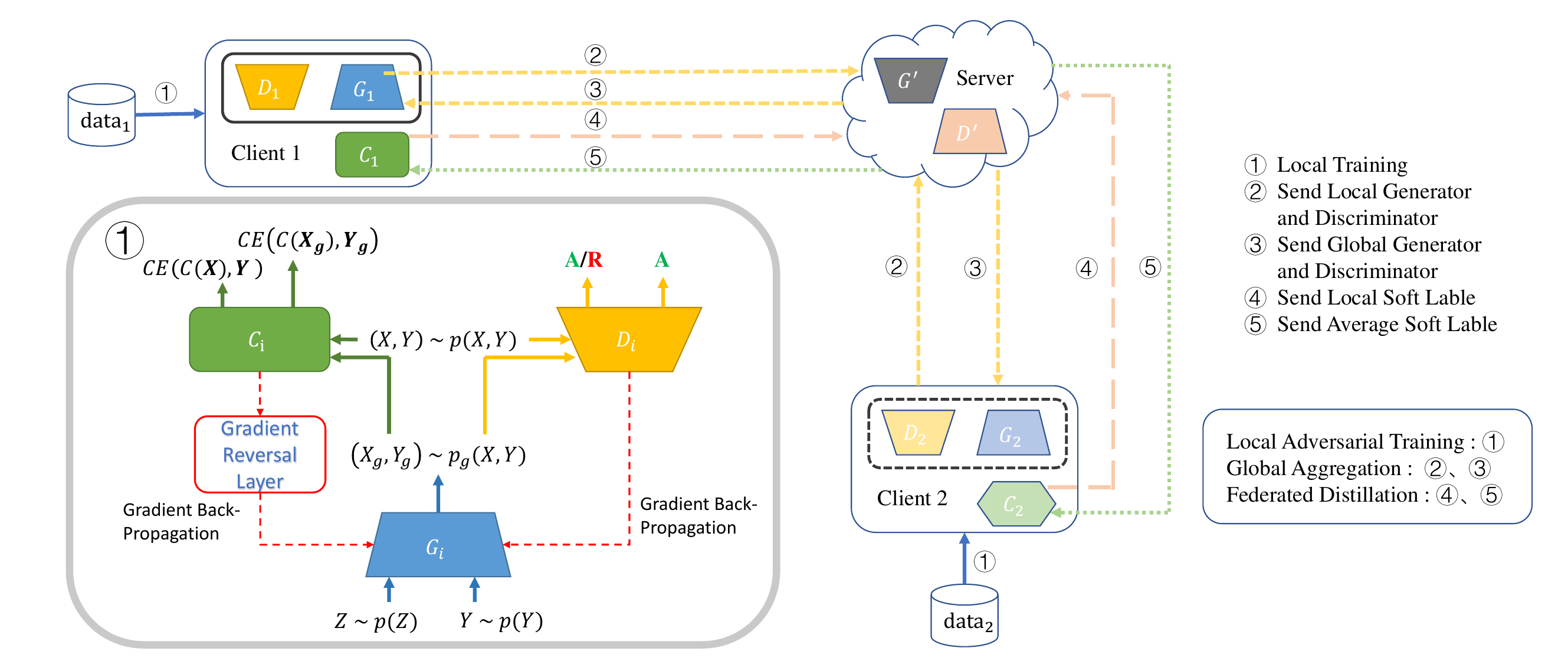}
    \caption{An illustration of FedDTG. Different colors represent different client model parameters, and different shapes represent different client model structures. For Local Adversarial Training, $i$ denotes the client index, 'R' denotes rejection, 'A' denotes acceptance, and 'CE' denotes the cross entropy loss.}
\label{f1}
\vspace{-0.15cm}
\end{figure}

\subsection{Local Adversarial Training}
    % \label{method1}
A standard GAN consists of a min-max optimization game, played between the generator $\mathbf{G}$ and discriminator $\mathbf{D}$. The objective function for this game is shown in Equation \ref{equ:GAN}: 
    \begin{equation}
    \begin{aligned}
        \displaystyle
        \min\limits_{\mathbf{G}} \max\limits_{\mathbf{D}} V(\mathbf{G},\mathbf{D})= & \mathbb{E}_{x\sim{p_{data}}}[\log{\mathbf{D}(x)} + \mathbb{E}_{z\sim{p_z}}[\log(1-\mathbf{D}(\mathbf{G}(z)))].
    \end{aligned}
    \label{equ:GAN}
    \end{equation}
    
Based on this, we introduce a local classifier $\mathbf{C}_k$ to the training process, forming a three-player GAN. This classifier, parameterized by $\theta_c^k$, aims to minimize the loss of the personalized client task. On the one hand, $\mathbf{C}_k$ can help accelerate $\mathbf{G}_k$ learn the local data distribution $p_{data}^k$; on the other hand, $\mathbf{G}_k$ assists $\mathbf{C}_k$ in alleviate the model drift caused by data heterogeneity during the local training phase.

Similar to the traditional GAN, $\mathbf{D}_k$ is a binary classifier that distinguishes real samples as true and generated samples as false. $\mathbf{D}_k$ needs to maximize the following objective function:
    \begin{equation}
    \begin{aligned}
        \displaystyle
        \mathcal{L}_{\mathbf{D}_k} = & \frac{1}{\Vert{data_k}\Vert}[\sum_{x\sim{p_{data_k}}} log \mathbf{D}_k(x) 
        \\+ &\sum_{z\sim p_z(z),\hat{y}\sim p_{\hat{y}}(\hat{y})} log (1-\mathbf{D}_k(\mathbf{G}(z,\hat{y})))].
    \end{aligned}
    \label{equ:D}
    \end{equation}
    
$\mathbf{G}_k$  takes noise vector $z\sim \mathcal N(0,1)$ and label vector $\hat{y} \sim \mathcal U(1,n)$ as inputs and outputs fake samples. The generated fake sample needs to be correctly classified by $\mathbf{C}_k$ and classified as true by the $\mathbf{D}_k$. With the help of $\mathbf{C}_k$, $\mathbf{G}_k$ better align the generated samples with the input labels and accelerate the learning of the local data distribution. In other words, the generator's goal is to produce samples that the discriminator cannot differentiate from real data, while also ensuring that the classifier correctly classifies the generated samples. $\mathbf{G}_k$ needs to minimize the following objective function:
    \begin{equation}
    \begin{aligned}
        \displaystyle
        \mathcal{L}_{\mathbf{G}_k} = &\frac{1}{\Vert{data_k}\Vert}\sum_{z\sim p_z(z),\hat{y}\sim p_{\hat{y}}(\hat{y})} CE(\mathbf{C}(\mathbf{G}(z,\hat{y})),\hat{y})
        \\ + &\sum_{z\sim p_z(z),\hat{y}\sim p_{\hat{y}}(\hat{y})} [1-log (\mathbf{D}_k(\mathbf{G}(z,\hat{y})))],
    \end{aligned}
    \label{equ:G}
    \end{equation}
where CE stands for cross-entropy loss function. $\mathbf{C}_k$ must accurately classify both the local data and the fake samples generated by $\mathbf{G}_k$. With the support of $\mathbf{G}_k$, $\mathbf{C}_k$ can avoid overfitting to the local training dataset, thereby reducing the risk of model drift. $\mathbf{C}_k$ needs to minimize the following objective function:
    \begin{equation}
    \begin{aligned}
        \displaystyle
        \mathcal{L}_{\mathbf{C}_k} = \frac{1}{\Vert{data_k}\Vert}[&\sum_{(x,y)\sim p_{data_k}}CE(\mathbf{C}_k(x),y)
        \\+&\sum_{(x_g,\hat{y})\sim{p_{\mathbf{G}_k}}}CE(\mathbf{C}_k(x_g),\hat{y})].
    \end{aligned}
    \label{equ:C}
    \end{equation}
 
Our local three-player GAN training is similar to \cite{vandenhende2019three}. However, their generator is designed to produce hard samples to accelerate the classifier's training, while our $\mathbf{G}_k$ is employed to help the $\mathbf{C}_k$ learn the global objective function better.

\subsection{Global Aggregation}
After local training, each client participating in this communication round is required to send $\mathbf{G}_k$ and $\mathbf{D}_k$ to the server. Then, the server uses parameter averaging to aggregate the global $\mathbf{G}$ and $\mathbf{D}$, parameterized by $\mathit{\theta_g}$ and $\mathit{\theta_d}$ which can be calculated by the following functions:
    \begin{equation}
    \begin{aligned}
        \displaystyle
        \mathit{\theta_g} = \sum_{k=1}^N \frac{1}{N} \mathit{\theta_g^k} , \mathit{\theta_d} = \sum_{k=1}^N \frac{1}{N} \mathit{\theta_d^k}.
    \end{aligned}
    \label{equ4}
    \end{equation}
    
It is worth noting that the local generator $\mathbf{G}_k$ does not directly access the real local dataset during the entire training process. Instead, the generated dataset is determined by noise vector $z\sim \mathcal N(0,1)$ and label vector $\hat{y} \sim \mathcal U(1,n)$, which is different from the local label distribution. For $\mathbf{D}_k$, it's a binary classifier and can not provide the probability information of each class. So it is unable to get the local label distribution from the uploaded $\mathbf{G}_k$ and $\mathbf{D}_k$. This approach effectively safeguards the privacy of local data. Finally, the server sends the updated global $\mathbf{G}$ and $\mathbf{D}$ back to the client.

At this stage, the new global $\mathbf{G}$ can obtain the outputs of all local classifiers $\mathbf{C}_k$ on the generated fake samples. We attempted to utilize these outputs to further train the global $\mathbf{G}$ in the server. This approach can potentially decrease the number of communication rounds and achieve better performance. To simplify the process, we only use the parameter averaging method to aggregate models, which can already achieve good results.

\subsection{Federated Distillation}
Upon receiving the models from the server, the client will replace the local $\mathbf{G}_k$ and $\mathbf{D}_k$ with the global $\mathbf{G}$ and $\mathbf{D}$.
The server then specifies the batch size and noise vector $\hat{z}\sim \mathcal N(0,1)$ for the federal distillation stage. For each batch, the client generates the same number of label vectors $\hat{y}$ for each class and inputs them into the $\mathbf{G}_k$ together with $\hat{z}$. Since the inputs and model parameters are consistent, all clients participating in this communication round outputs the same fake samples.
In this way, the communication cost caused by transmitting a large amount of generated data is avoided.

Input the fake samples into $\mathbf{C}_k$ to get the local soft labels, and send them to the server. The server then computes the average soft label of all clients except the current one and sends the result $y_{dis}^k$ which represents the ensemble knowledge back. Finally, the client computes the KL-divergence between the output of $\mathbf{C}_k$ and $y_{dis}^k$ as the distillation loss, and the loss of classifying the input fake samples. This allows $\mathbf{C}_k$ to learn from the knowledge of other clients. So, $\mathbf{C}_k$ needs to minimize the following objective function at this stage:
    \begin{equation}
    \begin{aligned}
            \displaystyle
            \mathcal{L}_{dis} = \frac{1}{\Vert{data_{dis}}\Vert}[\sum_{(x_g,\hat{y})\sim{p_{\mathbf{G}_k}}} {\alpha} KL(\mathbf{C}_k(x_g),y_{dis}^k)
            \\ +\sum_{(x_g,\hat{y})\sim{p_{\mathbf{G}_k}}} CE(\mathbf{C}_k(x_g),\hat{y})],
    \end{aligned}
    \label{equ:dis}
    \end{equation}
where $\alpha$ is the hyperparameter controlling the proportion of knowledge distillation loss. Here, we prefer to choose a large $\alpha$ value, because the classification generated fake samples has been included in the process of local adversarial training. The training procedure is shown in Algorithm\ref{alg:algorithm}
    \begin{algorithm}[H]
        \caption{FedDTG}
        \label{alg:algorithm}
        \textbf{Input}: local model $\mathbf{G}_k,\mathbf{D}_k,\mathbf{C}_k$\\
        % \textbf{Parameter}: Optional list of parameters\\
        \textbf{Output}: global $\mathbf{G}$ and N local $\mathbf{C}_k$
        \begin{algorithmic}[1] %[1] enables line numbers
        \STATE initialization local model parameters;
        \FOR{each communication round t = 1,...,T}
        \STATE \textbf{Local Adversarial Training}:
        \STATE $S_t \gets$ random subset(frac = 50\%) of the N clients
        \FOR{each client $k\in S_t$ \textbf{in parallel}}
        % \STATE $\theta_g^k \gets \theta_g,\theta_d^k \gets \theta_d$
        \FOR{n steps}
        \STATE calculate the loss $\mathcal{L}_{\mathbf{G}_k},\mathcal{L}_{\mathbf{D}_k},\mathcal{L}_{\mathbf{C}_k}$
        \STATE $\theta_g^k \gets \theta_g^k-\nabla \mathcal{L}_{\mathbf{G}_k}$,\\
        $\theta_d^k \gets \theta_d^k-\nabla \mathcal{L}_{\mathbf{D}_k}$,\\ 
        $\theta_c^k \gets \theta_c^k-\nabla \mathcal{L}_{\mathbf{C}_k}$
        \ENDFOR
        \STATE send $\theta_g^k$ and $\theta_d^k$ to serve
        \ENDFOR
        \STATE \textbf{Global Aggregation:}
        \STATE Server updates $\theta_g \gets \frac{1}{|S_t|}\sum_{k \in S_t} \theta_g^k,\theta_d \gets \frac{1}{|S_t|}\sum_{k \in S_t} \theta_d^k$
        and specify noise vector $\hat{z}\sim \mathcal N(0,1)$
        \STATE \textbf{Federated Distillation:}
        \FOR{each client $k\in S_t$ \textbf{in parallel}}
        \STATE $\theta_g^k \gets \theta_g,\theta_d^k \gets \theta_d$
        \STATE generate label vector $\hat{y}\sim \mathcal U(1,n)$
        \STATE input $\hat{y}$ and $\hat{z}$ to $\mathbf{G}_k$ to generate fake samples $x_g$
        \STATE input $x_g$ to $\mathbf{C}_k$ to get soft label $y_c^k$
        \STATE send $y_c^k$ to the serve.
        \ENDFOR
        \STATE Server calculates the average soft label $y_{dis}^k = \frac{1}{|S_t|}\sum_{i \in S_t}^{i\neq k} y_c^i$ and send it to each client\\
        \FOR{each client $k\in S_t$ \textbf{in parallel}}
        \STATE calculate the distillation loss $L_{dis}$
        \STATE $\theta_c^k \gets \theta_c^k-\nabla \mathcal{L}_{dis}$
        \ENDFOR
        \ENDFOR
        \end{algorithmic}
    \end{algorithm}

\section{Experiments}
\subsection{Setup}
\textbf{Dataset:} We experiment with competing approaches on three classic vision datasets for deep learning: MNIST \cite{lecun1998mnist}, EMNIST \cite{cohen2017emnist} and FashionMNIST \cite{xiao2017fashion}. MNIST dataset is composed of 28 x 28 pixels grayscale images with 6,000 training images and 1,000 testing images per digit. EMNIST dataset is for character image classification of the same size with 145,600 samples in total. FashionMNIST contains different clothing types, whose picture sizes, number of training and testing samples, and number of classes are exactly the same as MNIST.

\noindent \textbf{Baselines:} Our \emph{FedDTG} is designed to ensure safer model aggregation and provide a more effective solution for non-iid settings.
Thus, in addition to \emph{FedAvg} \cite{fedavg}, we compare FedDTG with methods for stabilizing local training and effective model aggregation: 
\emph{FedProx} \cite{fedprox} introduces a proximal term to constrain local model updates, thereby improving local training performance in heterogeneous environments.
\emph{FedDF} \cite{feddf} is designed for effective model fusion using a data-based knowledge distillation approach, where we provide a random 10\% of the training samples without labels as the proxy dataset.
\emph{FedGen} \cite{fedgen} is a federated distillation method that is data-free, incorporating flexible parameter sharing.

\noindent \textbf{Federated Learning Configurations:} Unless otherwise mentioned, we simulate 20 clients in total with an active-user ratio $frac=50\%$ each round. We use a Dirichlet distribution $\textbf{Dir}(\alpha)$ to simulate non-iid data distribution, which follows prior work \cite{li2021federated}. In the Dirichlet distribution, the value of $\alpha$ represents the degree of non-iid. 
\begin{table*}[]
    \centering
    \caption{Performance overview given different data settings. The best methods in each setting are highlighted in \textbf{bold} fonts.}
    \begin{tabular}{clccccc}
    \toprule
    \multicolumn{7}{c}{Average Test Accuracy} \\
    \midrule
    \multicolumn{1}{c}{Dataset} & Setting & \multicolumn{1}{c}{FedAvg} & \multicolumn{1}{c}{FedProx} & \multicolumn{1}{c}{FedDF} & \multicolumn{1}{c}{FedGen} & \multicolumn{1}{c}{\textbf{FedDTG}} \\
    \midrule
    \multirow{3}[2]{*}{\shortstack{MNIST\\$r=25\%$}} 
        & $\alpha$ =0.05 &$87.19\pm1.55$    &$88.64\pm1.42$ &$88.63\pm0.70$&$91.67\pm0.87$&$\mathbf{94.97\pm0.98}$\\
        & $\alpha$ =0.1  &$89.84\pm0.47$    &$89.79\pm0.48$ &$89.82\pm0.33$&$93.11\pm0.42$&$\mathbf{95.77\pm0.43}$\\
        & $\alpha$ =0.4  &$92.38\pm0.26$    &$92.83\pm0.14$ &$93.17\pm0.11$&$94.85\pm0.25$&$\mathbf{96.02\pm0.11}$  \\
    \midrule
    \multirow{3}[2]{*}{\shortstack{MNIST\\$r=10\%$}}
        & $\alpha$ =0.05 &$85.86\pm2.46$    &$87.38\pm2.37$ &$88.69\pm0.64$&$90.44\pm0.44$&$\mathbf{93.89\pm0.99}$\\
        & $\alpha$ =0.1  &$87.19\pm1.49$    &$88.63\pm1.07$  &$88.86\pm0.51$&$92.88\pm0.19$&$\mathbf{95.02\pm0.47}$\\
        & $\alpha$ =0.4  &$91.27\pm0.28$    &$91.52\pm0.26$ & $92.33\pm0.17$&$93.87\pm0.13$&$\mathbf{95.27\pm0.24}$  \\
    \midrule
    \multirow{3}[2]{*}{\shortstack{EMNIST\\$r=10\%$}} 
        & $\alpha$ =0.05 &$62.43\pm1.07$    &$61.92\pm1.48$ &$63.57\pm1.14$&$66.26\pm1.36$&$\mathbf{73.50\pm0.20}$\\
        & $\alpha$ =0.1  &$66.70\pm0.96$    &$65.62\pm0.72$ & $67.01\pm0.87$&$70.84\pm0.67$&$\mathbf{74.44\pm0.28}$  \\
        & $\alpha$ =0.4  &$71.11\pm0.32$    &$71.36\pm0.31$ &$72.68\pm0.37$&$75.31\pm0.52$&$\mathbf{75.46\pm0.19}$ \\
    \midrule
    \multirow{3}[2]{*}{\shortstack{Fashion\\MNIST\\$r=10\%$}} 
        & $\alpha$ =0.05 &$68.68\pm2.12$    &$69.60\pm1.52$ &$66.92\pm0.81$&$69.40\pm1.03$&$\mathbf{77.94\pm0.32}$\\
        & $\alpha$ =0.1  &$71.98\pm1.04$    &$71.99\pm1.01$ &$69.15\pm0.92$&$73.86\pm0.33$&$\mathbf{79.90\pm0.29}$  \\
        & $\alpha$ =0.4  &$78.90\pm0.67$    &$78.87\pm0.66$ &$78.51\pm0.38$&$78.58\pm0.36$&$\mathbf{80.31\pm0.24}$ \\
    \bottomrule
    \end{tabular}%
    % \vspace{-0.15cm}
    \label{tab:2}%
\end{table*}% 
The smaller $\alpha$, the higher the degree of data heterogeneity, which is different in size, class distributions, and the distribution of each class. 
We use three values of $\alpha$ = 0.05, 0.1, 0.4 to simulate the setting of non-iid.
For each local training, we use the Adam optimizer with a learning rate = 0.0001, epoch = 20, batch size = 32, and a five-layer convolution network for each client.
We use 10000 generated images each round for FedDTG to do federated distillation. For fair comparison and reflecting the federal environment lacking local data, we use at most $r=25\%$ training datasets and distribute them to the clients to simulate the non-iid federated setting and use all testing datasets for performance evaluation.

\subsection{Results}
\noindent \textbf{Impacts of data heterogeneity:} As shown in Table \ref{tab:2}, FedDTG achieves optimal and stable results under different levels of data heterogeneity.
It is worth noting that FedProx and FedDF are optimized based on the average of participant model parameters by FedAvg, and FedGen also uses part of the parameter sharing.  
However, in FedDTG, the knowledge of other models is transferred in the mutual distillation between clients, and there is no direct parameter averaging
of the classifier model. So, this task is more difficult than other algorithms. Our FedDTG still performs better compared with other methods, especially when the value of $\alpha$ is small. This thoroughly validates the outstanding performance of our method in addressing data heterogeneity issues. In addition, the reduction of training data has minimal impact on FedDTG. 

This result is in line with our motivation. With the help of distributed GAN, 
the generated fake samples not only expand the local training data 
but also limit the local objective function from excessively deviating from the global objective function.
The use of soft labels on fake samples for knowledge distillation will further alleviate the problem of heterogeneous data distribution. This type of knowledge is otherwise not accessible by baselines such as FedAvg and FedProx.

\begin{figure}[]
    \centering
    \includegraphics[width=0.48\linewidth]{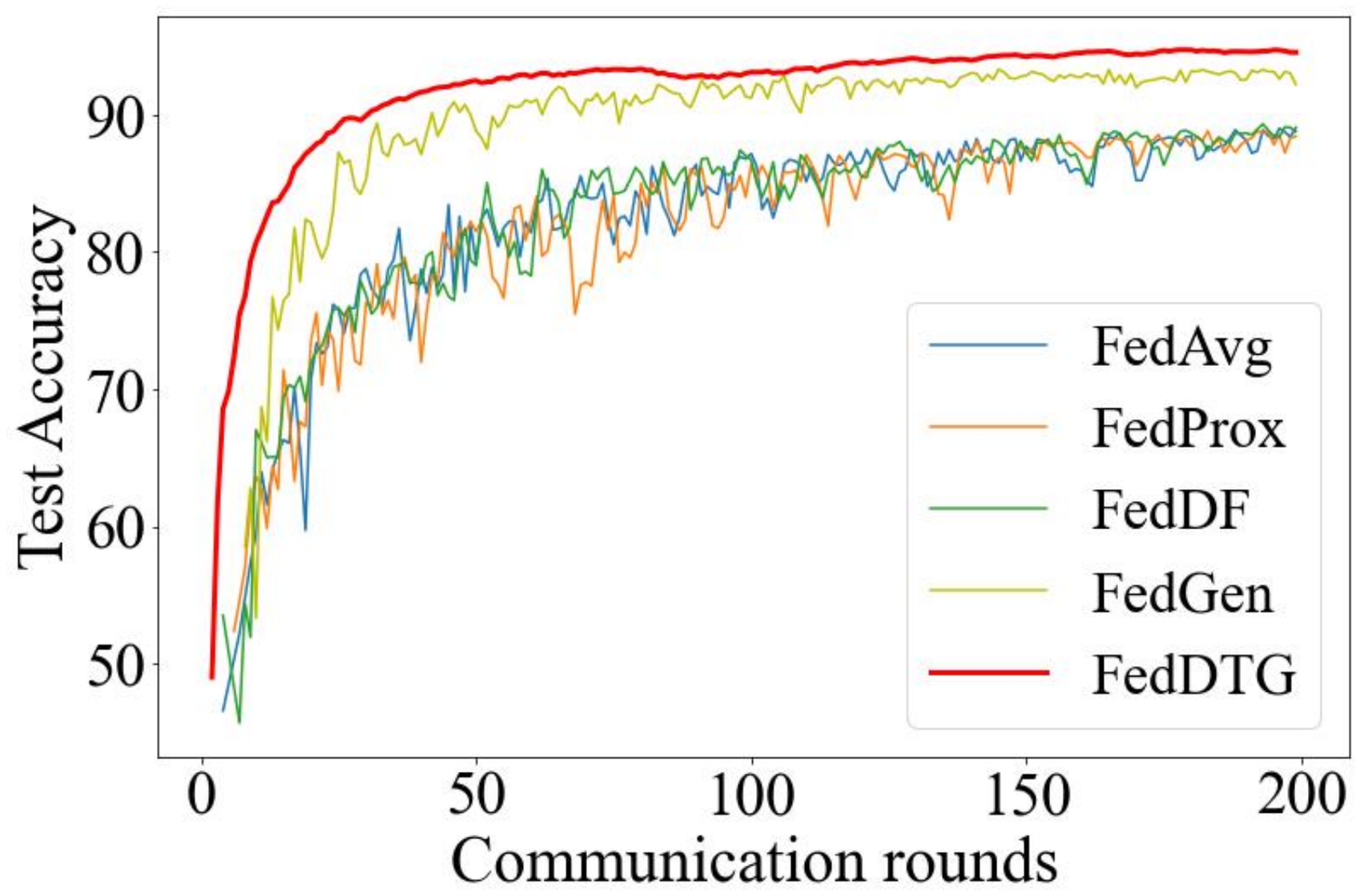}
    \includegraphics[width=0.48\linewidth]{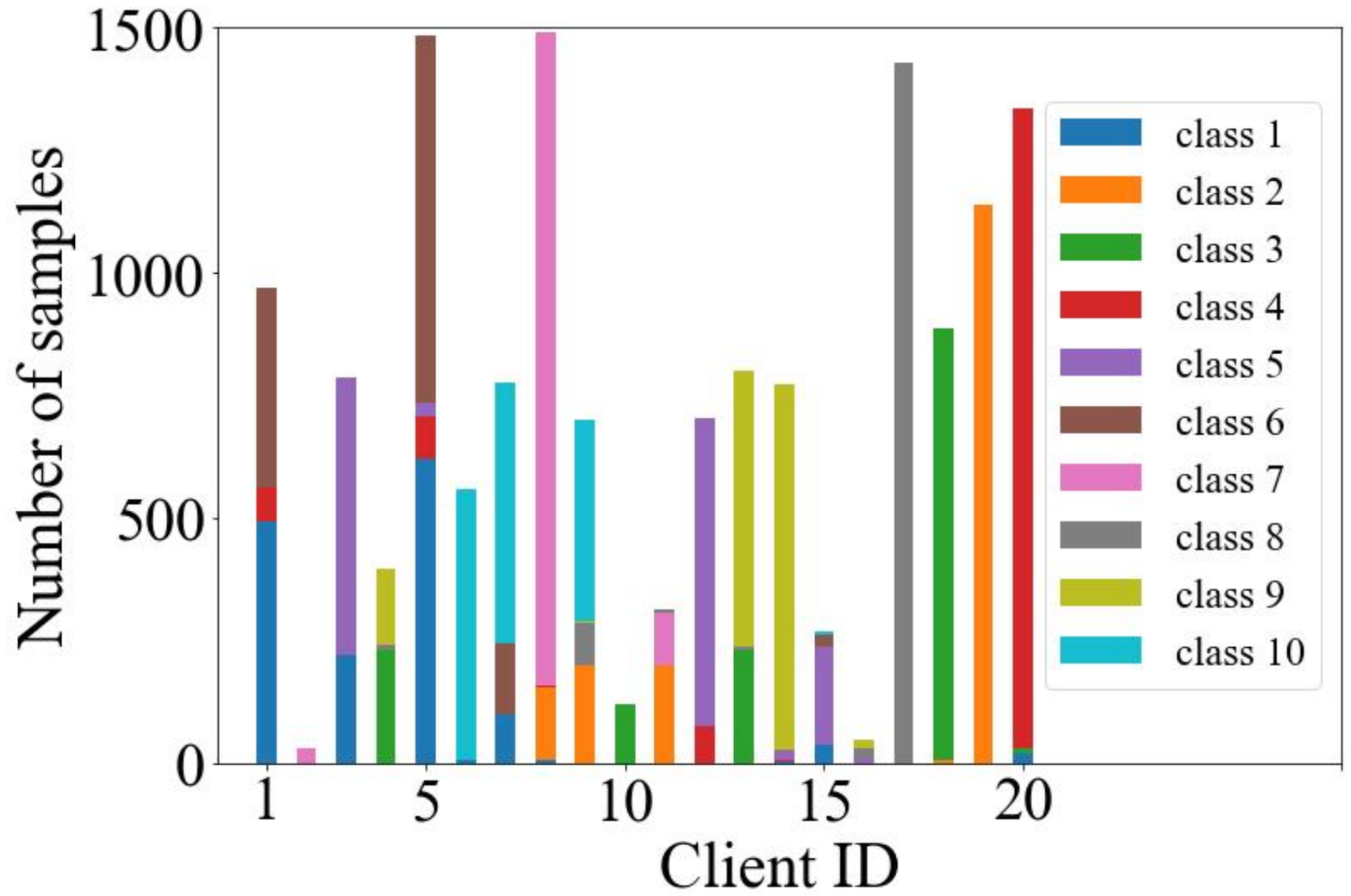}
    \caption{Performance on MNIST under $\alpha=0.05,r=25\%$ and Illustration of the number of samples per class allocated to each client.}
\vspace{-0.15cm}
\label{figure3}
\end{figure}

FedDF is distilled on the global model obtained by FedAvg, so with the increase of the degree of non-iid, it will have a certain effect, but the improvement is not great. 
Different from FedDF, the performance gain of our FedDTG is significant compared with FedAvg. This discrepancy shows that our proposed approach of directly doing mutual distillation between the client models is more effective than the method of fine-tuning the global model which is based on FedAvg.

As one of the most competitive baselines, FedGen achieves good results in most cases, but it does not make full use of the method of knowledge distillation. 
As we can see, our proposed approach outperformance FedGen in most cases, especially when the degree of data heterogeneity is high. In FedDTG, the local client model not only needs to correctly classify the generated images but also needs to distill the knowledge of other clients, which is more stable than FedGen completely relying on the generator to transmit information.

% From the perspective of the dataset, we found that the accuracy of all methods on MNIST is much higher than that of EMNIST and Fashion MNIST. We believe that the difference in performance of all methods is mainly due to the different complexity of the datasets. EMNIST is more complex than MNIST because it contains more categories and greater sample variability. Fashion MNIST contains images of clothing from different categories, and its visual features are more complex, making it more difficult for the model to distinguish between these images. However, our method has achieved the best performance on both simple and complex datasets, which further illustrates the application potential of our method under various feature data.
\begin{figure}[h]
    \centering
    \includegraphics[width=0.48\linewidth]{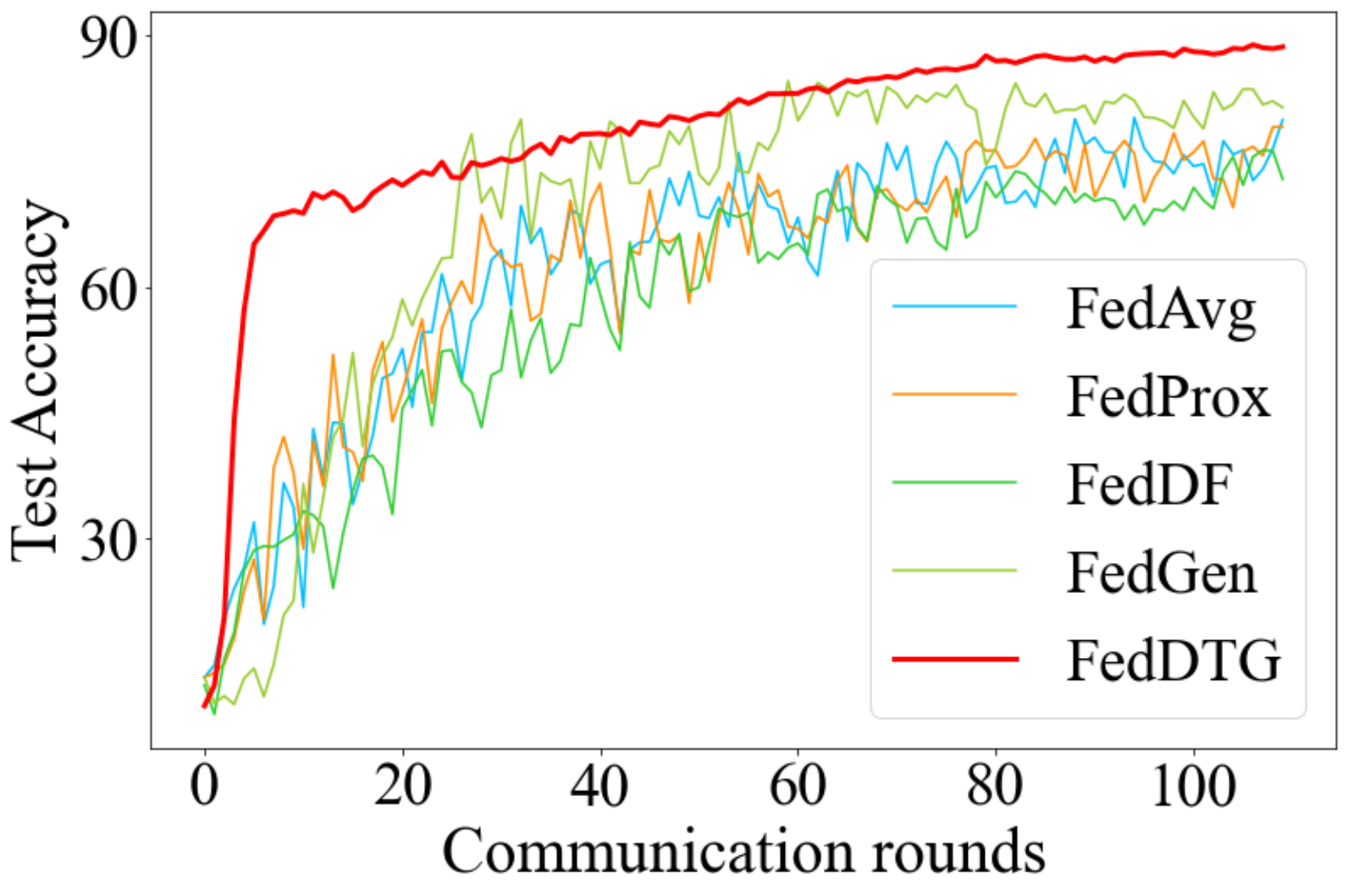}
    \includegraphics[width=0.48\linewidth]{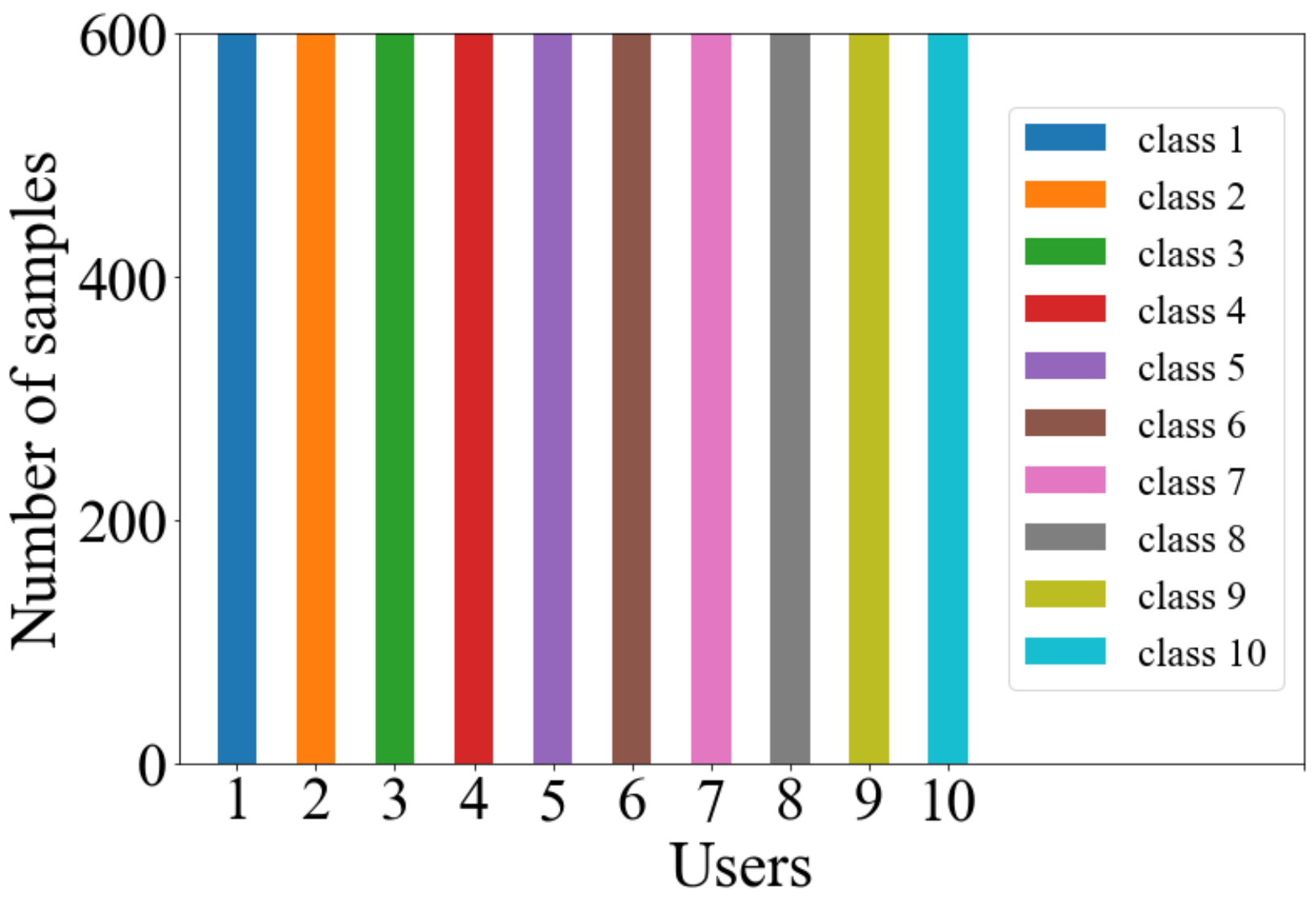}
    \vfill
    \includegraphics[width=0.48\linewidth]{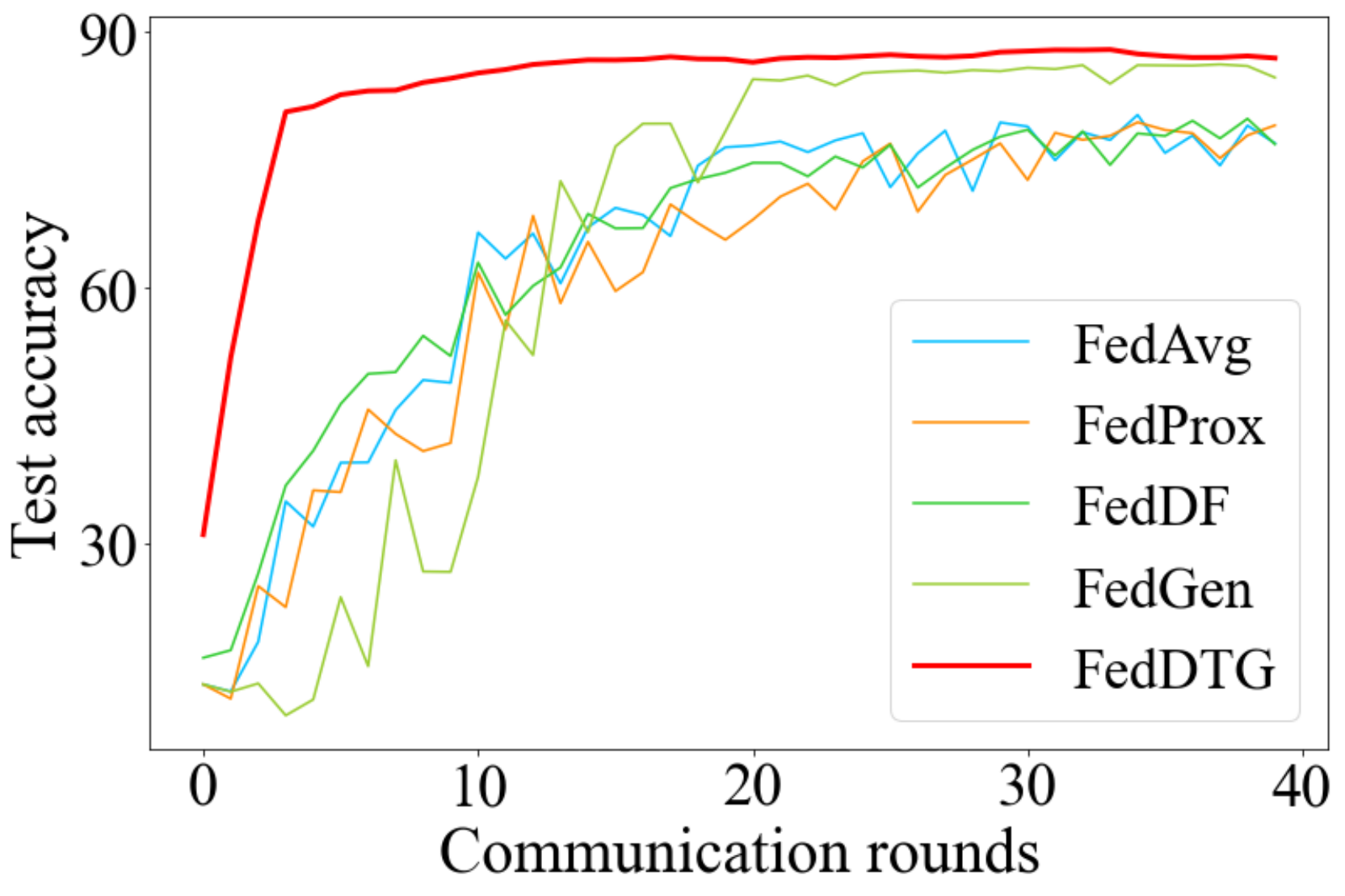}
    \includegraphics[width=0.48\linewidth]{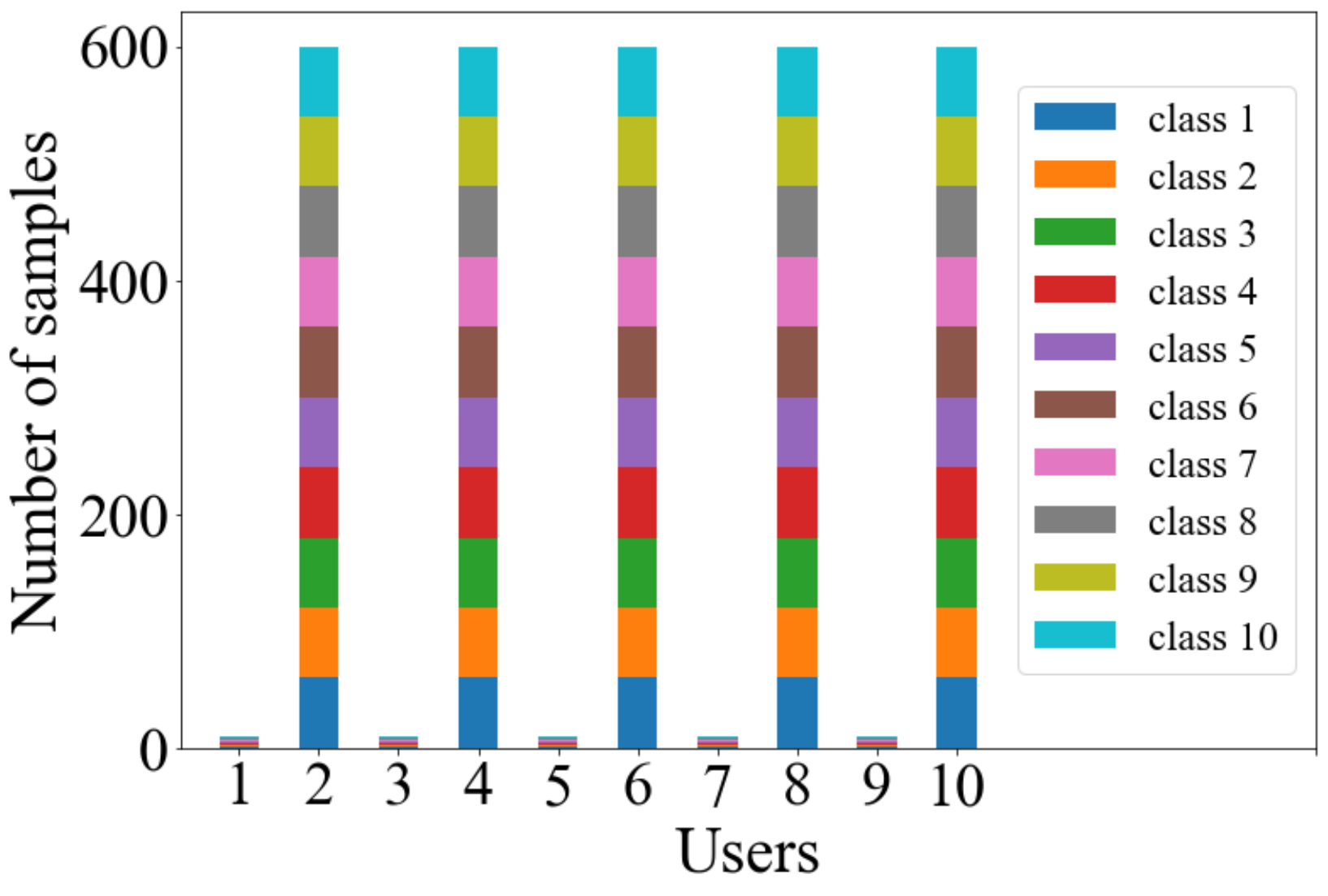}
\caption{Performance on two extremely unbalanced label distribution and quantity distribution.}
\label{figure6}
\vspace{-0.15cm}
\end{figure}

\noindent \textbf{Learning efficiency:}
Since FedDTG does not have a global classifier, we use the average accuracy of all client models on the test dataset as the final result. 
As we can see in Figure \ref{figure3}, FedDTG has the most rapid and stable learning curves and outperformance other baselines on MNIST under $\alpha=0.05,r=25\%$. We show the number of samples per class allocated to each client under this federated setting on the right side of Figure \ref{figure3}. We can see that in this extreme non-iid case, most clients have only a single class of images, and some clients even have very few local training data. This greatly increases the difficulty of federated training. Although FedGen can also achieve high learning efficiency under certain data settings, our approach has less fluctuation in accuracy during training.

\noindent \textbf{Robustness:}
We tested the generalization performance of FedDTG under two extremely unbalanced label distributions and quantity distributions. 
We set 10 clients and $r=10\%$ training data on the MNIST dataset. In the first case, each client has only one class of data. In the second case, half of the clients have very little image data, and the other half have relatively much image data, where each client has the same number of data belonging to each class. As shown in Figure \ref{figure6}, the stability of FedDTG has obvious advantages over other algorithms, which once again verifies our approach can perform well in various situations. Furthermore, the mutual distillation between clients accelerates their access to global knowledge, helping the quicker convergence of the classifier model. This shows to some extent that our method can reduce the number of communication rounds between the client and the server and has faster efficiency.

\begin{figure}[h]
\begin{minipage}{0.5\linewidth}
    \centerline{\includegraphics[width=0.95\linewidth]{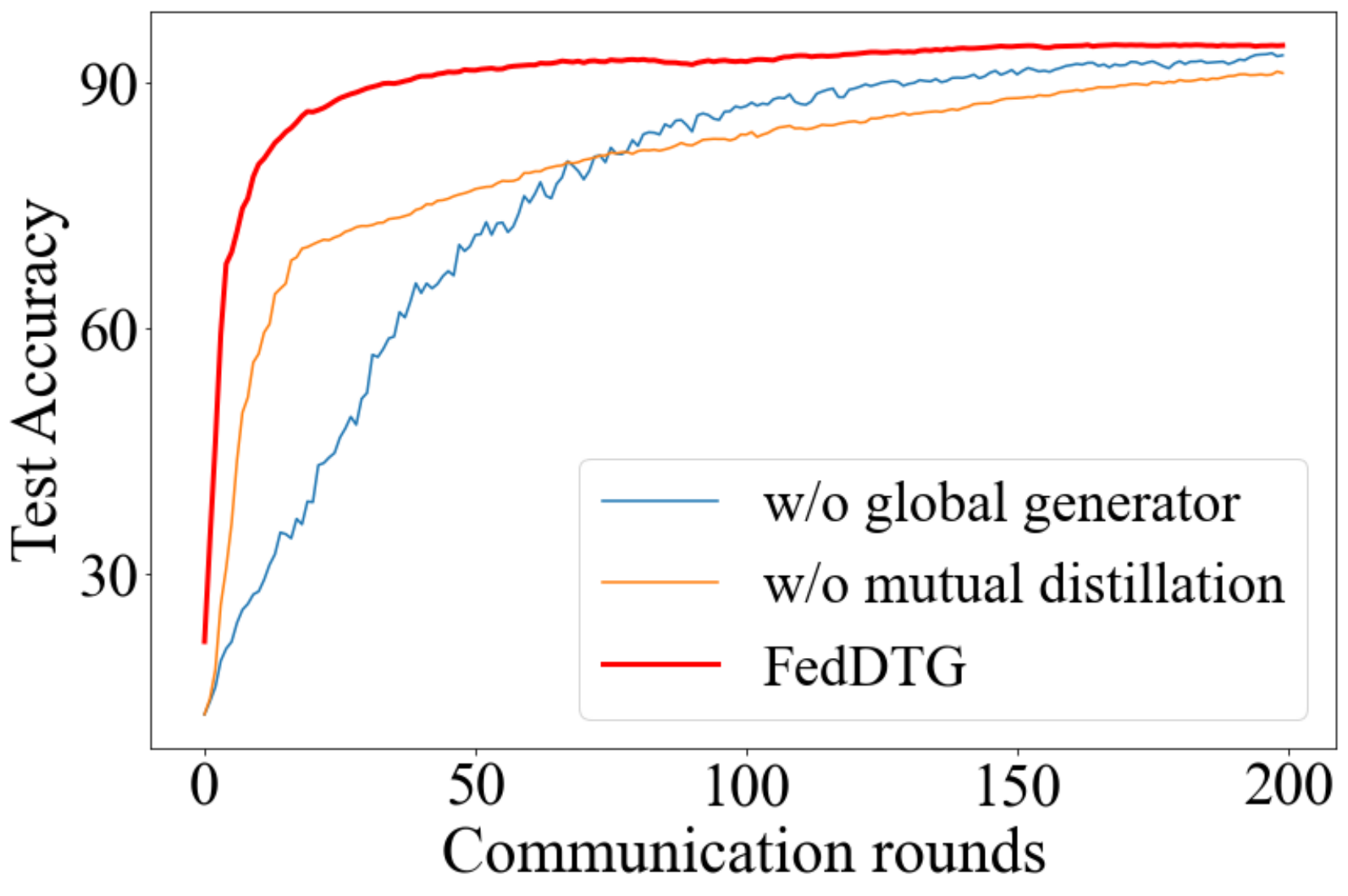}}
\end{minipage}
% \hfill 
\begin{minipage}{0.48\linewidth}
    \centering
    \begin{tabular}{cc}
        \begin{tabular}{cc}
            \toprule& \textbf{Test acc} \\
            \midrule
            local training & 29.3 \\
            \midrule
            \multirow{2}[2]{*}{\shortstack{w/o mutual distillation}} & \multirow{2}[2]{*}{92.33} \\&  \\
            \midrule
            \multirow{2}[2]{*}{\shortstack{w/o global generator}} & \multirow{2}[2]{*}{93.78} \\&  \\
            \midrule
            \textbf{FedDTG} & \textbf{94.97} \\
            \bottomrule
            \end{tabular}%
    \end{tabular}
\end{minipage}
\caption{Ablation study of FedDTG. The dataset is MNIST under $\alpha=0.05,r=25\%$.}
\vspace{-0.35cm}
\label{abl}
\end{figure}

\subsection{Ablation study}
We tested the effects of different parts of FedDTG on the final experimental results.
The transmission of information between clients in FedDTG mainly comes from the global $\mathbf{G}$ and the mutual distillation between clients.
If mutual distillation is used alone, the whole method is similar to the traditional federated distillation after removing the global teacher model. If the global generator $\mathbf{G}$ is used alone, it is equivalent to directly expanding the local dataset with the generated images. As shown on the left side of Figure \ref{abl}, 
the global generator is capable of producing a large number of synthetic samples, expanding the data available for client training and reducing the accuracy fluctuation. Through mutual distillation, each model can learn from the knowledge of other clients without directly accessing their data after each training iteration. This approach enhances the model's capabilities and further accelerates convergence.
Both of them have significantly improved the results.

\subsection{Personalization performance}
\begin{figure}[]
    \centering
    \includegraphics[width=1.0\textwidth]{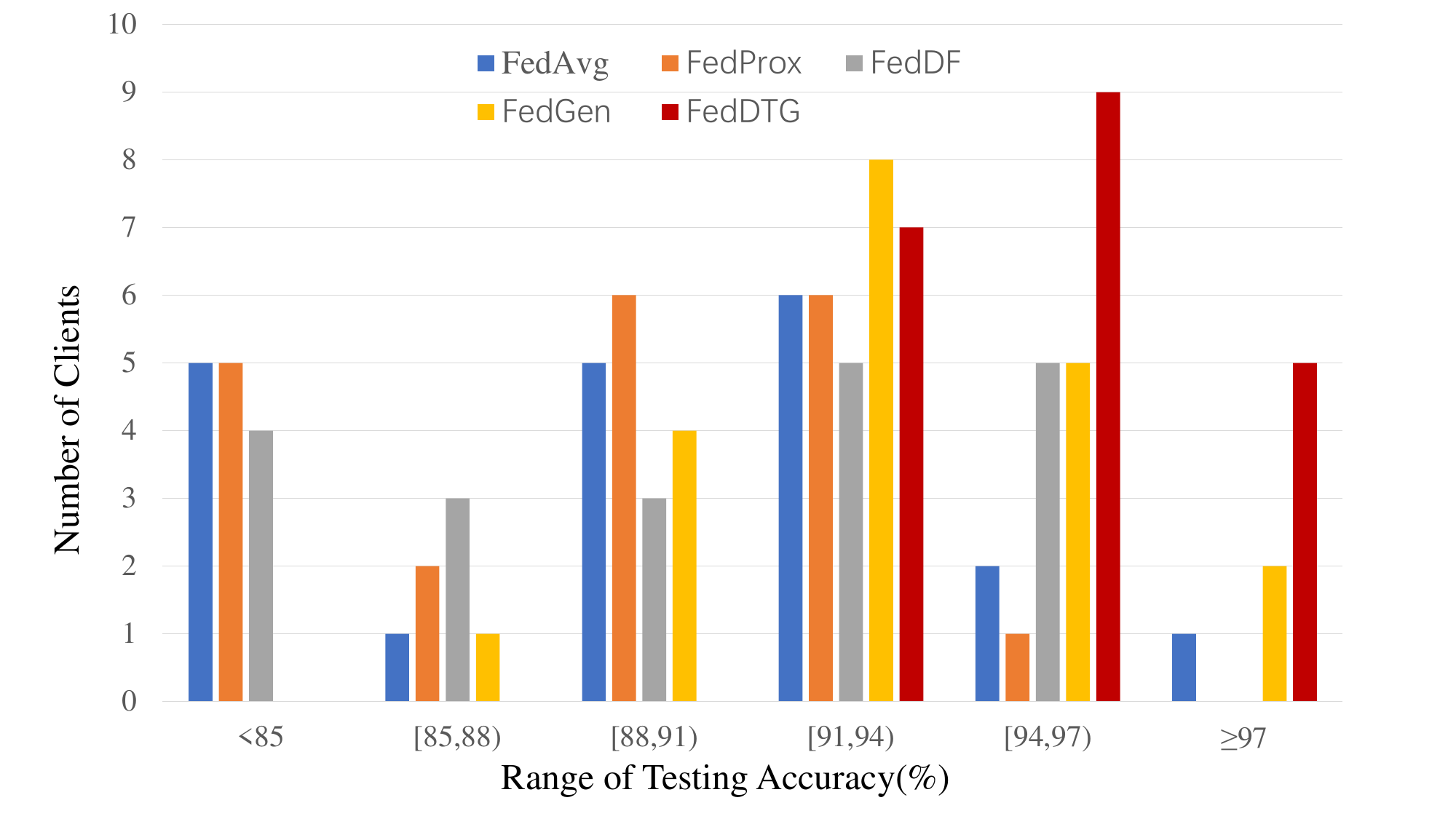}
    \caption{The distribution of the testing accuracy of all clients with MNIST under $\alpha=0.05,r=25\%$.}
\vspace{-0.15cm}
\label{figure5}
\end{figure}

% \noindent \textbf{Personalization performance:}\\
To evaluate the personalization performance of all methods in detail, we also distribute the test datasets to clients with the same $\textbf{Dir}(\alpha=0.05)$ for local performance evaluation. We analyze the test accuracy of the personalized model owned by each client in Figure \ref{figure5}. For FedDTG, even the worst client achieves an accuracy of 92.15\% on the local test dataset which is a huge improvement compared with other algorithms. In addition, the test accuracy of all clients in FedDTG is relatively centralized, which once again proves the advantages of the global generator and mutual distillation.

\section{Conclusion and Future Work}
In this paper, we address the challenging problem of federated learning where clients are reluctant to share their personalized models and propose 
FedDTG. Specifically, we introduce a novel distributed GAN to help federated mutual distillation between clients without compromising their data privacy. As validated by extensive experiments, our method can significantly improve performance more rapidly and with greater stability. 

\paragraph{\textbf{Acknowledgments.}} This preprint has no post-submission improvements or corrections. The Version of Record of this contribution is published in the Neural Information Processing, ICONIP 2024 Proceedings and is available online at https://doi.org/10.1007/978-981-96-6969-1\_2.
% As technology develops, GAN has been extended to many variants, and the effect is improving. In the future, we can apply more abundant GANs to our method and conduct experiments on more complex and diverse datasets. We will further explore our method in more fields such as finance, so that our method can be directly applied to more real-world scenarios.

% \paragraph{\textbf{Acknowledgments.}} This work was supported by the National Key Research and Development Project of China (2021YFC3340300), the Zhejiang Provincial Key Research and Development Project (2023C01043), and Academy Of Social Governance Zhejiang University.





% LNCS Homepage, \url{http://www.springer.com/lncs}, last accessed 2023/10/25
% \end{thebibliography}

\bibliographystyle{splncs04}
\bibliography{0412}

\begin{thebibliography}{10}
\providecommand{\url}[1]{\texttt{#1}}
\providecommand{\urlprefix}{URL }
\providecommand{\doi}[1]{https://doi.org/#1}

\bibitem{cohen2017emnist}
Cohen, G., Afshar, S., Tapson, J., Van~Schaik, A.: Emnist: Extending mnist to handwritten letters. In: 2017 International Joint Conference on Neural Networks (IJCNN). pp. 2921--2926. IEEE (2017)

\bibitem{dwork2006differential}
Dwork, C.: Differential privacy. In: International colloquium on automata, languages, and programming. pp. 1--12. Springer (2006)

\bibitem{g3}
Fan, C., Liu, P.: Federated generative adversarial learning. In: Chinese Conference on Pattern Recognition and Computer Vision (PRCV). pp. 3--15. Springer (2020)

\bibitem{goodfellow2014generative}
Goodfellow, I., Pouget-Abadie, J., Mirza, M., Xu, B., Warde-Farley, D., Ozair, S., Courville, A., Bengio, Y.: Generative adversarial nets. Advances in neural information processing systems  \textbf{27} (2014)

\bibitem{mdgan}
Hardy, C., Le~Merrer, E., Sericola, B.: Md-gan: Multi-discriminator generative adversarial networks for distributed datasets. In: 2019 IEEE international parallel and distributed processing symposium (IPDPS). pp. 866--877. IEEE (2019)

\bibitem{hinton2015distilling}
Hinton, G., Vinyals, O., Dean, J.: Distilling the knowledge in a neural network. arXiv preprint arXiv:1503.02531  (2015)

\bibitem{huang2021personalized}
Huang, Y., Chu, L., Zhou, Z., Wang, L., Liu, J., Pei, J., Zhang, Y.: Personalized cross-silo federated learning on non-iid data. In: Proceedings of the AAAI Conference on Artificial Intelligence. vol.~35, pp. 7865--7873 (2021)

\bibitem{Fedsurvey}
Kairouz, P., McMahan, H.B., Avent, B., Bellet, A., Bennis, M., Bhagoji, A.N., Bonawitz, K., Charles, Z., Cormode, G., Cummings, R., et~al.: Advances and open problems in federated learning. Foundations and Trends{\textregistered} in Machine Learning  \textbf{14}(1--2),  1--210 (2021)

\bibitem{scaffold}
Karimireddy, S.P., Kale, S., Mohri, M., Reddi, S., Stich, S., Suresh, A.T.: Scaffold: Stochastic controlled averaging for federated learning. In: International Conference on Machine Learning. pp. 5132--5143. PMLR (2020)

\bibitem{lecun1998mnist}
LeCun, Y.: The mnist database of handwritten digits. http://yann. lecun. com/exdb/mnist/  (1998)

\bibitem{fedmd}
Li, D., Wang, J.: Fedmd: Heterogenous federated learning via model distillation. arXiv preprint arXiv:1910.03581  (2019)

\bibitem{li2021federated}
Li, Q., Diao, Y., Chen, Q., He, B.: Federated learning on non-iid data silos: An experimental study. arXiv preprint arXiv:2102.02079  (2021)

\bibitem{li2021model}
Li, Q., He, B., Song, D.: Model-contrastive federated learning. In: Proceedings of the IEEE/CVF Conference on Computer Vision and Pattern Recognition. pp. 10713--10722 (2021)

\bibitem{fedprox}
Li, T., Sahu, A.K., Zaheer, M., Sanjabi, M., Talwalkar, A., Smith, V.: Federated optimization in heterogeneous networks. Proceedings of Machine Learning and Systems  \textbf{2},  429--450 (2020)

\bibitem{noniid}
Li, X., Huang, K., Yang, W., Wang, S., Zhang, Z.: On the convergence of fedavg on non-iid data. arXiv preprint arXiv:1907.02189  (2019)

\bibitem{feddf}
Lin, T., Kong, L., Stich, S.U., Jaggi, M.: Ensemble distillation for robust model fusion in federated learning. arXiv preprint arXiv:2006.07242  (2020)

\bibitem{long2020federated}
Long, G., Tan, Y., Jiang, J., Zhang, C.: Federated learning for open banking. In: Federated Learning: Privacy and Incentive, pp. 240--254. Springer (2020)

\bibitem{fedavg}
McMahan, B., Moore, E., Ramage, D., Hampson, S., y~Arcas, B.A.: Communication-efficient learning of deep networks from decentralized data. In: Artificial intelligence and statistics. pp. 1273--1282. PMLR (2017)

\bibitem{g1}
Rasouli, M., Sun, T., Rajagopal, R.: Fedgan: Federated generative adversarial networks for distributed data. arXiv preprint arXiv:2006.07228  (2020)

\bibitem{sheller2020federated}
Sheller, M.J., Edwards, B., Reina, G.A., Martin, J., Pati, S., Kotrotsou, A., Milchenko, M., Xu, W., Marcus, D., Colen, R.R., et~al.: Federated learning in medicine: facilitating multi-institutional collaborations without sharing patient data. Scientific reports  \textbf{10}(1),  12598 (2020)

\bibitem{shen2020federated}
Shen, T., Zhang, J., Jia, X., Zhang, F., Huang, G., Zhou, P., Kuang, K., Wu, F., Wu, C.: Federated mutual learning. arXiv preprint arXiv:2006.16765  (2020)

\bibitem{vandenhende2019three}
Vandenhende, S., De~Brabandere, B., Neven, D., Van~Gool, L.: A three-player gan: generating hard samples to improve classification networks. In: 2019 16th International Conference on Machine Vision Applications (MVA). pp.~1--6. IEEE (2019)

\bibitem{xiao2017fashion}
Xiao, H., Rasul, K., Vollgraf, R.: Fashion-mnist: a novel image dataset for benchmarking machine learning algorithms. arXiv preprint arXiv:1708.07747  (2017)

\bibitem{yin2021see}
Yin, H., Mallya, A., Vahdat, A., Alvarez, J.M., Kautz, J., Molchanov, P.: See through gradients: Image batch recovery via gradinversion. In: Proceedings of the IEEE/CVF Conference on Computer Vision and Pattern Recognition. pp. 16337--16346 (2021)

\bibitem{yonetani2019decentralized}
Yonetani, R., Takahashi, T., Hashimoto, A., Ushiku, Y.: Decentralized learning of generative adversarial networks from non-iid data. arXiv preprint arXiv:1905.09684  (2019)

\bibitem{zhang2018deep}
Zhang, Y., Xiang, T., Hospedales, T.M., Lu, H.: Deep mutual learning. In: Proceedings of the IEEE Conference on Computer Vision and Pattern Recognition. pp. 4320--4328 (2018)

\bibitem{g2}
Zhou, X., Liu, X., Lan, G., Wu, J.: Federated conditional generative adversarial nets imputation method for air quality missing data. Knowledge-Based Systems  \textbf{228},  107261 (2021)

\bibitem{fedgen}
Zhu, Z., Hong, J., Zhou, J.: Data-free knowledge distillation for heterogeneous federated learning. arXiv preprint arXiv:2105.10056  (2021)

\end{thebibliography}

\end{document}